\documentclass[oneside,english,reqno]{amsart}

\usepackage{algorithm}
\usepackage{algorithmic}
\usepackage{multirow}
\usepackage[top=2cm,bottom=2.5cm]{geometry}
\usepackage{fancyhdr} 
\usepackage{enumitem}

\usepackage[T1]{fontenc}
\usepackage[utf8]{inputenc}
\usepackage{babel}
\usepackage{amsmath}
\usepackage{amsthm}
\usepackage{amssymb}

\usepackage{comment}

\usepackage{graphicx}
\usepackage{tikz}
\usetikzlibrary{shapes.misc}
\usepackage{float}

\usepackage[labelformat=simple]{subfig}

\usepackage[unicode=true,pdfusetitle,
bookmarks=true,bookmarksnumbered=false,bookmarksopen=false,
breaklinks=false,pdfborder={0 0 1},backref=false,hidelinks]
{hyperref}

\makeatletter

\numberwithin{equation}{section}
\numberwithin{figure}{section}

\makeatother

\begin{document}

\title[Octree in point cloud analysis]{The use of Octree in point cloud analysis with application to cultural heritage
}

\keywords{octree, 3D point cloud, data classification}
\author{
	Rafał Bieńkowski$^{1}$
}
\author{
	 Krzysztof Rutkowski$^2$
}
\thanks{$^1$ Systems Research Institute of the Polish Academy of Sciences, 01-447 Warsaw, Poland, Newelska 6, 		
}

\thanks{$^2$ Cardinal Stefan Wyszy\'nski University, 
	Faculty of Mathematics and Natural Sciences. School of Exact Sciences,
	Warsaw, Poland, Dewajtis 5}

\begin{abstract}In this article we present the effects of our work on the subject of the technical approach to the 3D point cloud data analysis through the use of the Octree method to compress, analyse and compute the initial data.
\end{abstract}
\maketitle\thispagestyle{empty}

\section{Introduction}

3D documentation and renderings are becoming more and more ubiquitous in numerous fields, such as engineering, architecture, urban planning, large-scale landscape analysis, and cultural heritage (Art History, Archaeology, and Museum Studies). With the ongoing improvement of acquisition tools (e.g. laser scanning, photogrammetry, LiDAR) and methods of 3D model generation (3D modelling and prototyping software), the accuracy and resolution of widely available 3D data have greatly improved. 

In our article, we address two aspects of handling large 3D point clouds, that is size reduction and point classification. For both of these aspects, we apply the octree approach. 

The process of improving the 3D data quality follows a similar development to the use of 2D images, from small bitmaps to high-resolution images. As in the 2D case, for the purpose of storage, analysis or transfer 3D files should be reduced in size without any significant loss of quality. 

For a 3D point cloud to be useful, in most applications, the points need to be classified first. In many applications, a large number of points can be classified as noise. 
Below, we propose an approach to size reduction in 3D point clouds. We focus on the detection of two types of areas present in point clouds: 1) vegetation, and 2) regions of insufficient point density to produce reliable documentation.

Below, we propose one such approach to size reduction of 3D data coming from the field of Archaeology. 

\section{State of the art}
Topography point cloud analysis is a time and resource consuming process, especially in terms of manual analysis, like classification. There are a lot of different methods of point cloud creation such as laser scanning or Structure from Motion (SfM) \cite{MARKIEWICZ2019224}. In the case of our experiment, we use the database on the SfM method, based on collecting 2D images and computing them to create a 3D object.

The idea of Octree was first published by Donald Meagher in 1980 as a method to represent and process 3D objects in computer graphics \cite{Meagher}. In modern scientific work, there are a lot of publications on the application of Octree in different fields of computer science. Below we would like to mention just a couple of examples:
\begin{enumerate}
    \item Octree Grid Topology – used to segment objects that have a known topology \cite{Bai},
    \item Nearest neighbour search \cite{Drost2018},
    \item Colour Quantization \cite{articlePARK}.
\end{enumerate}

\section{Problem statement}

In the present contribution, we investigate the use of the octree method for the size reduction and classification of 3D point clouds. In the experiments and analysis, we use numerical data sets representing an area of cultural heritage interest (an archaeological trench, documented during ongoing fieldwork and its surroundings - topographical data). The data sets are given in the form of point clouds based on photogrammetry. Each point in the point cloud (data set) is represented by its georeferenced position in space and its colour is given in the RGB system.

In our investigation, we use the Octree method to choose points to be merged based on the distance criterion. In the 3D Octree method, a space/object is represented by cuboids, of various sizes. If points are “close enough” in cuboids of a suitable length of the edges, points are merged.

\section{Data sets}
\label{data}

Below we present a short description of the data sets used in the investigation. For the preliminary results, presented in Section “Numerical experiment”, we used three data sets. All sets come from the photogrammetric documentation (based on image processing) of an archaeological site. Photogrammetric documentation in our data sets has been created in Agisoft Metashape based on the orthogonal photos taken from a drone.

The sets are as follows: 

Set 1 -- a point cloud documenting a cross-section of an archaeological trench with remains of architecture (stone walls) inside the trench. The points cloud covers an area around 2,5 by 6 meters in plan and ca. 70-100 cm deep. 

Set 2 -- similar to Set 1, this set represents/documents part of an archaeological trench, but with a strong focus on its surroundings, not the contents of the trench. This point cloud covers an area of ca. 3,5 by 6,5 meters. During the acquisition of this data, the vegetation around the trench was also of interest, hence the vertical measurements registered on the cloud are from 2m above ground (tree height) to 1 m of depth (inside trench).

Set 3 -- a point cloud documenting a part of the archaeological heritage site. The points cloud covers an area around 10 by 10 meters and ca. 30 meters in height. This data set has been chosen based on the high vegetation in the centre of the documented area - a large tree. 

All points have location data in a georeferenced coordinates system. In our case, it is the UTM coordinate system with data represented as latitude, longitude and elevation. The UTM zone codes for our data sets are UTM37T and UTM38T. An example of the location data for one, the selected point takes the following form (case of variable sites in Georgia).

\section{Data processing}

It is characteristic for geographic data to reverse the order of the first two axes, namely the first values given are from $y$ axis, the second values are given $x$ axis and the third values given are from $z$ axis. 
In our application, we decided to work with the geographic order of the axis and therefore we used this order of data in our algorithm.

From the data set we extract the following information:
\\[-0.25cm]
\begin{itemize}
	\item $y_\text{min}$ – minimal value on y-axis of vertex,
	\item $y_\text{max}$ – maximal value on y-axis of vertex,
    \item $x_\text{min}$ – minimal value on x-axis of vertex,
	\item $x_\text{max}$ – maximal value on x-axis of vertex,
	\item $z_\text{min}$ – minimal value on z-axis of vertex,
	\item $z_\text{max}$ – maximal value on z-axis of vertex.
\end{itemize}
\mbox{\ }\\[-0.25cm]
and due to the very small differences of the $yx$-position of points (which differs on at most 4 positions after the decimal point), we perform the following change of the $yx$-data: for $y$-values and $x$-values we drop the decimal precision before 4 positions after the decimal point and scale the result by $10^4$. The value of $z$ remains unchanged.

The selected level cuboids, which contain points, are of dimensions
\begin{equation*}
\frac{y_\text{max}-y_\text{min}}{lev}  \times 
    \frac{x_\text{max}-x_\text{min}}{lev} \times         \frac{z_\text{max}-z_\text{min}}{lev},
\end{equation*}
where $lev$ represents the maximal level of division in the Octree method.\newpage
The Octree method is as follows

\begin{enumerate}
    \item 	Given data we put into a cuboid of dimension\\ $(y_\text{max}-y_\text{min}) \times
    (x_\text{max}-x_\text{min}) \times  (z_\text{max}-z_\text{min})$,
	\item If the cuboid contains a vertex, split the cuboid into 8 cuboids of equal dimensions (by dividing each edge by two),
	\item For each new cuboid we assign step 2, whenever the level of nesting is less or equal to lev.

\end{enumerate}
The preview of this procedure is illustrated in the following graph.\\[0cm]
\begin{center}
\begin{tikzpicture}[scale=0.4]
\coordinate (S1) at (0,0);
\coordinate (S2) at (4,0);
\coordinate (S4) at (0,4);
\coordinate (S3) at (4,4);
\coordinate (S21) at (5.4,0.7);
\coordinate (S41) at (1.4,4.7);
\coordinate (S31) at (5.4,4.7);
\draw (S1) -- (S2) -- (S3) -- (S4) -- (S1);
\draw (S2) -- (S21) -- (S31) -- (S3);
\draw (S31) -- (S41) -- (S4);
\draw[dashed,->] (5.75,2) -- (7.5,2); 
\begin{scope}[shift={(8,0)}]
\coordinate (S1) at (0,0);
\coordinate (S2) at (4,0);
\coordinate (S4) at (0,4);
\coordinate (S3) at (4,4);
\coordinate (S21) at (5.4,0.7);
\coordinate (S41) at (1.4,4.7);
\coordinate (S31) at (5.4,4.7);
\draw (S1) -- (S2) -- (S3) -- (S4) -- (S1);
\draw (S2) -- (S21) -- (S31) -- (S3);
\draw (S31) -- (S41) -- (S4);
\coordinate (U1) at (0,2);
\coordinate (U2) at (4,2);
\coordinate (U3) at (2,0);
\coordinate (U4) at (2,4);
\draw (U1) -- (U2);
\draw (U3) -- (U4);
\coordinate (R1) at (5.4,2.7);
\coordinate (R2) at (4.7,0.35);
\coordinate (R3) at (4.7,4.35);
\draw (U2) -- (R1); 
\draw (R2) -- (R3); 
\coordinate (W1) at (0.7,4+0.35);
\coordinate (W2) at (3.4,4.7);
\draw (W1) -- (R3);
\draw (W2) -- (U4);
\end{scope}
\draw[dashed,->] (13.75,2) -- (15.5,2); 
\begin{scope}[shift={(16,0)}]
\coordinate (S1) at (0,0);
\coordinate (S2) at (4,0);
\coordinate (S4) at (0,4);
\coordinate (S3) at (4,4);
\coordinate (S21) at (5.4,0.7);
\coordinate (S41) at (1.4,4.7);
\coordinate (S31) at (5.4,4.7);
\draw (S1) -- (S2) -- (S3) -- (S4) -- (S1);
\draw (S2) -- (S21) -- (S31) -- (S3);
\draw (S31) -- (S41) -- (S4);
\coordinate (U1) at (0,2);
\coordinate (U2) at (4,2);
\coordinate (U3) at (2,0);
\coordinate (U4) at (2,4);
\draw (U1) -- (U2);
\draw (U3) -- (U4);
\coordinate (R1) at (5.4,2.7);
\coordinate (R2) at (4.7,0.35);
\coordinate (R3) at (4.7,4.35);
\draw (U2) -- (R1); 
\draw (R2) -- (R3); 
\coordinate (W1) at (0.7,4+0.35);
\coordinate (W2) at (3.4,4.7);
\draw (W1) -- (R3);
\draw (W2) -- (U4);
\coordinate (UU1) at (3,2);
\coordinate (UU2) at (3,4);
\coordinate (UU3) at (2,3);
\coordinate (UU4) at (4,3);
\draw (UU1)-- (UU2);
\draw (UU3) -- (UU4);
\coordinate (RR1) at (4.7,3.35);
\coordinate (RR2) at (4.35,2.175);
\coordinate (RR3) at (4.35,4.175);
\draw (RR1) -- (UU4);
\draw (RR2)-- (RR3);
\coordinate (WW1) at (2+0.35,4.175);
\coordinate (WW2) at (3+0.7,4.35);
\draw (UU2) -- (WW2);
\draw (RR3) -- (WW1);
\end{scope}
\end{tikzpicture}
\end{center}\mbox{\ }\\[-0.5cm]

The result of this procedure are cuboids up to the desired level of nesting. From the cuboids of maximal depth, we extract the cuboids which contain vertices. 
In our approach, we have chosen to use cuboids. It is 
 however possible to implement a similar method using exclusively cubes by choosing initial data to be contained in a cube. We found that cuboids are better suited to our application. In a real-life application, the whole process benefits from the use of cuboids, instead of cubes, as cuboids are able to fit better into the investigated shapes.   

\section{Algorithm}
 The algorithm to classify the point cloud firstly sorts the cuboids of max depth which contain vertices with respect to $z_k$, $z_k=z_{\text{min}}+\frac{z_{\text{max}}-z_{\text{min}}}{lev}$, $k=0,1\dots,2^{lev}-1$ values for $y$,$x$ dimensions $[y_i,y_{i+1}]\times [x_{j},x_{j+1}]$, $i,j\in 0,\dots,2^{lev}-1$. Then for each coordinate $[y_i,y_{i+1}]\times [x_{j},x_{j+1}]$, $i,j\in 0,\dots,2^{lev}-1$ we find the connected cuboids of minimal height $z$. We mark these cuboids as "surface" and the rest cuboids of these coordinates mark as "above". 
 
 The whole process of processing the data is illustrated in the following graph:
\\[0cm]
\begin{center}
\begin{tikzpicture}[rotate=-90]
\coordinate (S1) at (0,0);
\node[draw,rounded corners=2,align=center] (N1) at (S1) {Input obj};
\coordinate (S2) at (1.5,0);
\node[draw,rounded corners=2,align=center] (N2) at (S2) {Data in matrix $n\times 3$};
\draw[->] (N1)-- node[right,align=left] {Read obj package\\ vertices positions} (N2); 
\coordinate (S3) at (3,0);
\node[draw,rounded corners=2,align=center] (N3) at (S3) {Octree structure};
\draw[->] (N2)-- node[right] {Built Octree package} (N3); \coordinate (S4) at (4.5,0);
\node[draw,rounded corners=2,align=center] (N4) at (S4) {Cuboids of selected level\\
containing vertices};
\draw[->] (N3)-- node[right] {Intersection}(N4); 
\coordinate (S5) at (6,0);
\node[draw,rounded corners=2,align=center] (N5) at (S5) {Classified cuboids};
\draw[->] (N4)--  node[right] {Algorithm} (N5); 
\end{tikzpicture}
\end{center}
\newpage
The algorithm is presented as follows:

\begin{algorithm}[H]
\caption{Finding surface cuboids}
\begin{algorithmic} 
\FOR{each cuboid of selected depth level containing vertices}
\STATE{Sort the cuboid data with respect to the $z$ variable for each $yx$-coordinate}
\ENDFOR
\FOR{For each $yx$ coordinate}
\FOR{For each following pair of cuboids in $yx$ cordinate\footnotemark
}
\IF{Distance between two following cuboids in $z$ is greater than $0$}
\STATE{Mark the first cuboid as "surface"}
\STATE{Break the loop of "For each cuboid of $yx$ coordinate"}
\ELSE
\STATE{Mark the first cuboid as "surface"}
\ENDIF
\ENDFOR
\ENDFOR
\end{algorithmic}
\end{algorithm}
\footnotetext{Note, that if there is only one cuboid in $yx$ coordinate then we mark it as "surface"}
The cost of the algorithm is $O(lev^3)$, since in the pessimistic case we need to analyse each of the $yxz$ cuboids of the selected level depth.  

Below we present the second algorithm to classify cuboids "surface" and "above" and also fill the gap cuboids between cuboids "surface"-"above" or "above"-"above" as "gap" cuboids.

\begin{algorithm}[H]
\caption{Finding surface, above cuboids and gaps}
\begin{algorithmic} 
\FOR{each cuboid of selected depth level containing vertices}
\STATE{Sort the cuboid data with respect to the $z$ variable for each $yx$-coordinate}
\ENDFOR
\FOR{For each $yx$ coordinate}
\FOR{For each following pair of cuboids in $yx$ cordinate}
\IF{Distance between two following cuboids in $z$ is greater than $0$}
\STATE{Mark the second cuboid as "above" and the following cuboids of $yx$ coordinates as "above"}
\STATE{Fill in the cuboid of coordinate $yx$, height $[z_1,z_{2}]$, where $z_1$ is the max height of the first cuboid of the pair, $z_2$ is the min height of the second cuboid of the pair}
\ENDIF
\ENDFOR
\ENDFOR
\STATE{Mark the cuboids of selected depth level containing vertices which are not "above" as "surface".}
\end{algorithmic}
\end{algorithm}

\section{Numerical Experiment}

We consider data from three sets, as described in Section \ref{data}. Sets $1$, $2$, $3$ are made up of $626831$,  $1219669$ and $993802$ points respectively. The level of depth of cuboids was set to $5$ (starting from the initial level $0$).

Each of the points has a representation in $x,y,z$ values of the georeferenced coordinates system and the colour in the RGB system. The numerical experiment was performed on the computer with the following hardware parameters: processor AMD Ryzen 9 3950X 16-Core, 128 GB RAM DDR4. The software used for calculations was MatLab 2020 with the help of the readObj package (see \cite{readobj}) and the OCTree package (see \cite{octree}. The result of performing Octree on the data by choosing the most nested cuboids, which contain points of the data is displayed in the following pictures:

For the respective data sets the number of cuboids is as follows:
\begin{itemize}
    \item For set 1 – 12297 cuboids on levels 0-5, where 5003 cuboids of level 5 contain vertices,
    \item For set 2 – 8281 cuboids on levels 0-5, where 3479 cuboids of level 5 contain vertices,
    \item For set 3 - 4792 cuboids on levels 0-5, where 1677 cuboids of level 5 contain vertices.
\end{itemize}

The results are displayed in the appendix A. Figures.

\bibliographystyle{plain}
\bibliography{bibliography}

\appendix
\cfoot{\thepage}
\pagestyle{fancy}
\section{Figures}
\begin{figure}[htbp]
\centering
\subfloat[]{\includegraphics[width=0.45\linewidth]{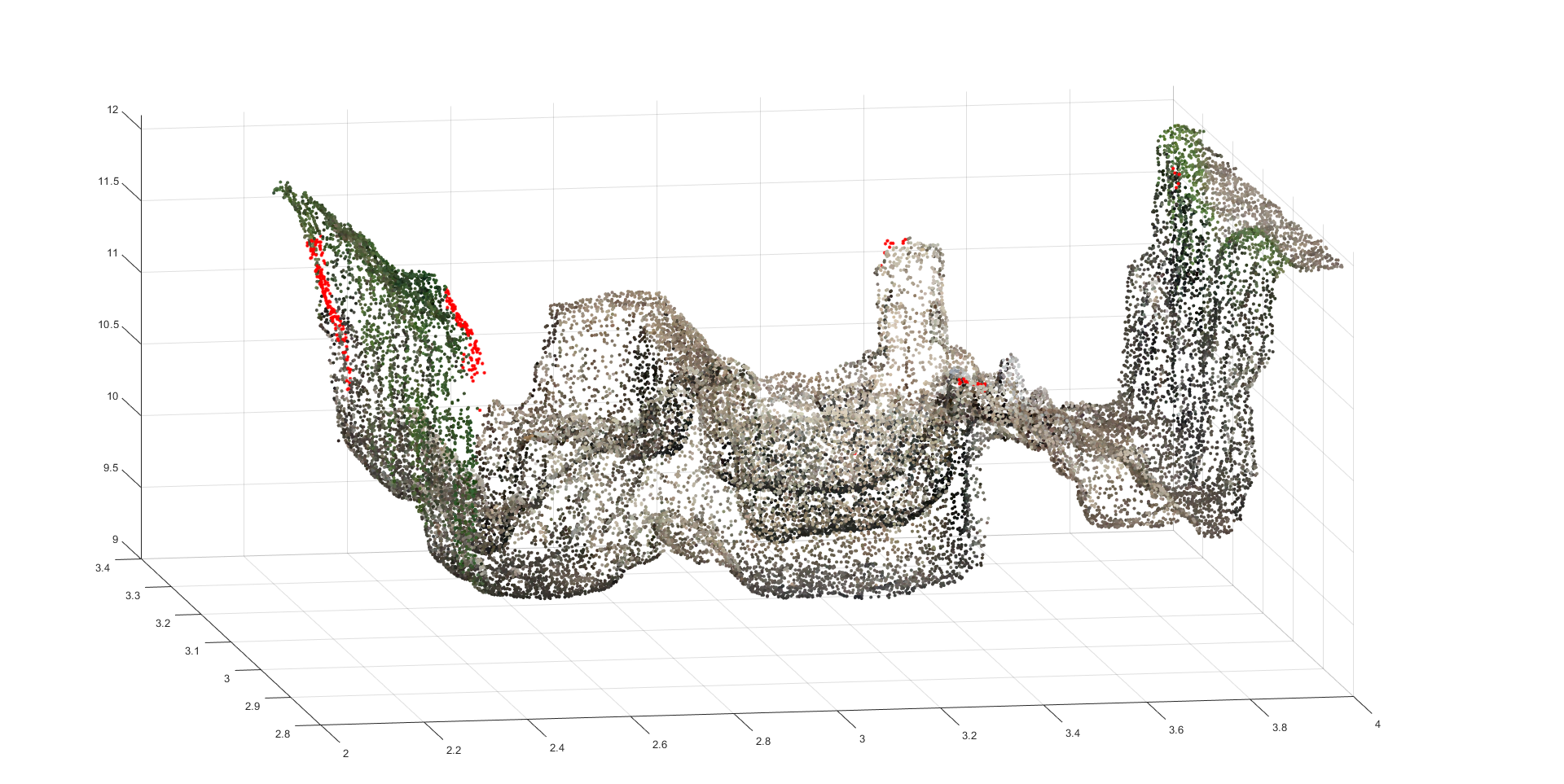}}\
\subfloat[]{\includegraphics[width=0.45\linewidth]{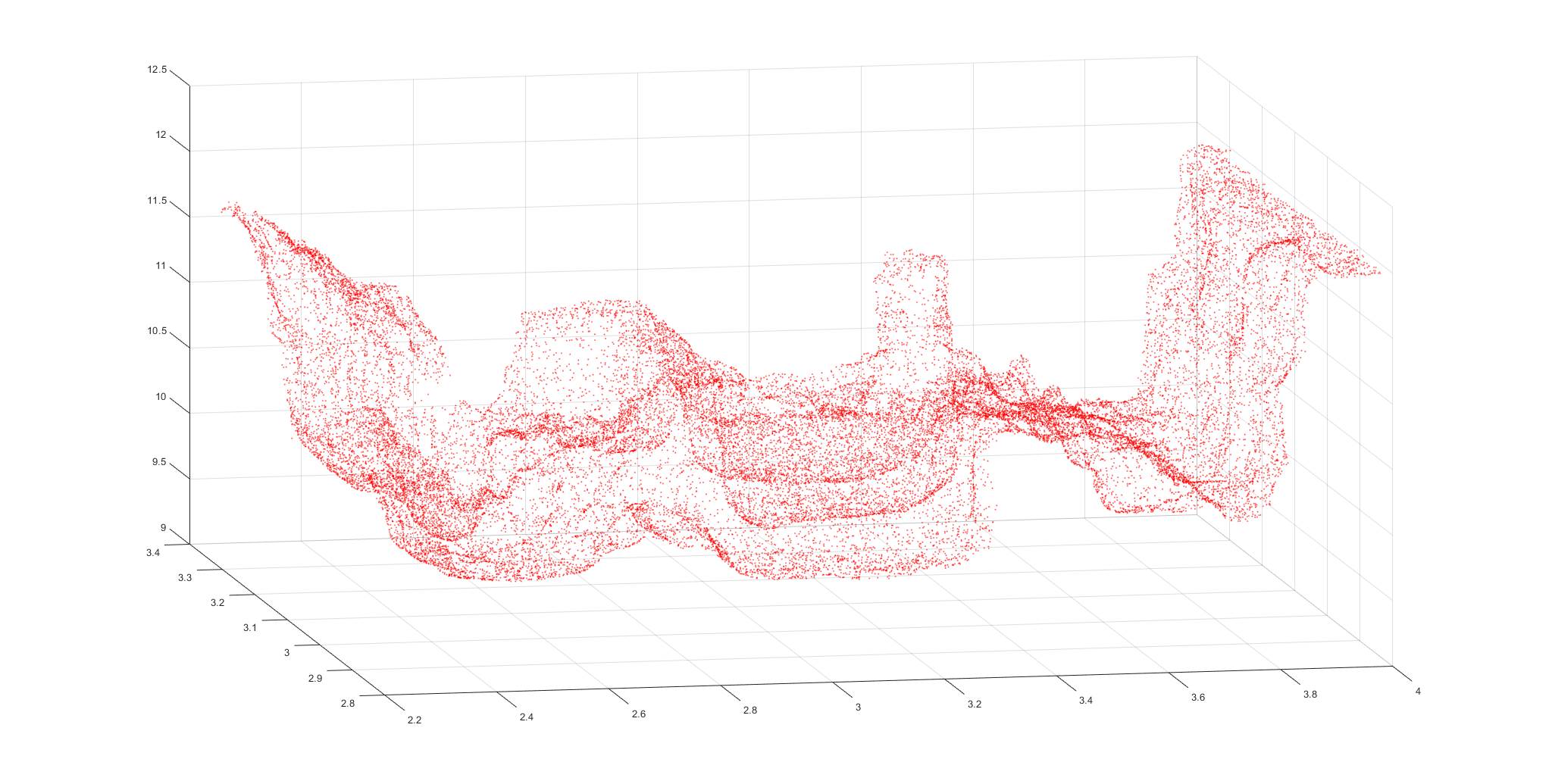}}\\
\subfloat[]{\includegraphics[width=0.45\linewidth]{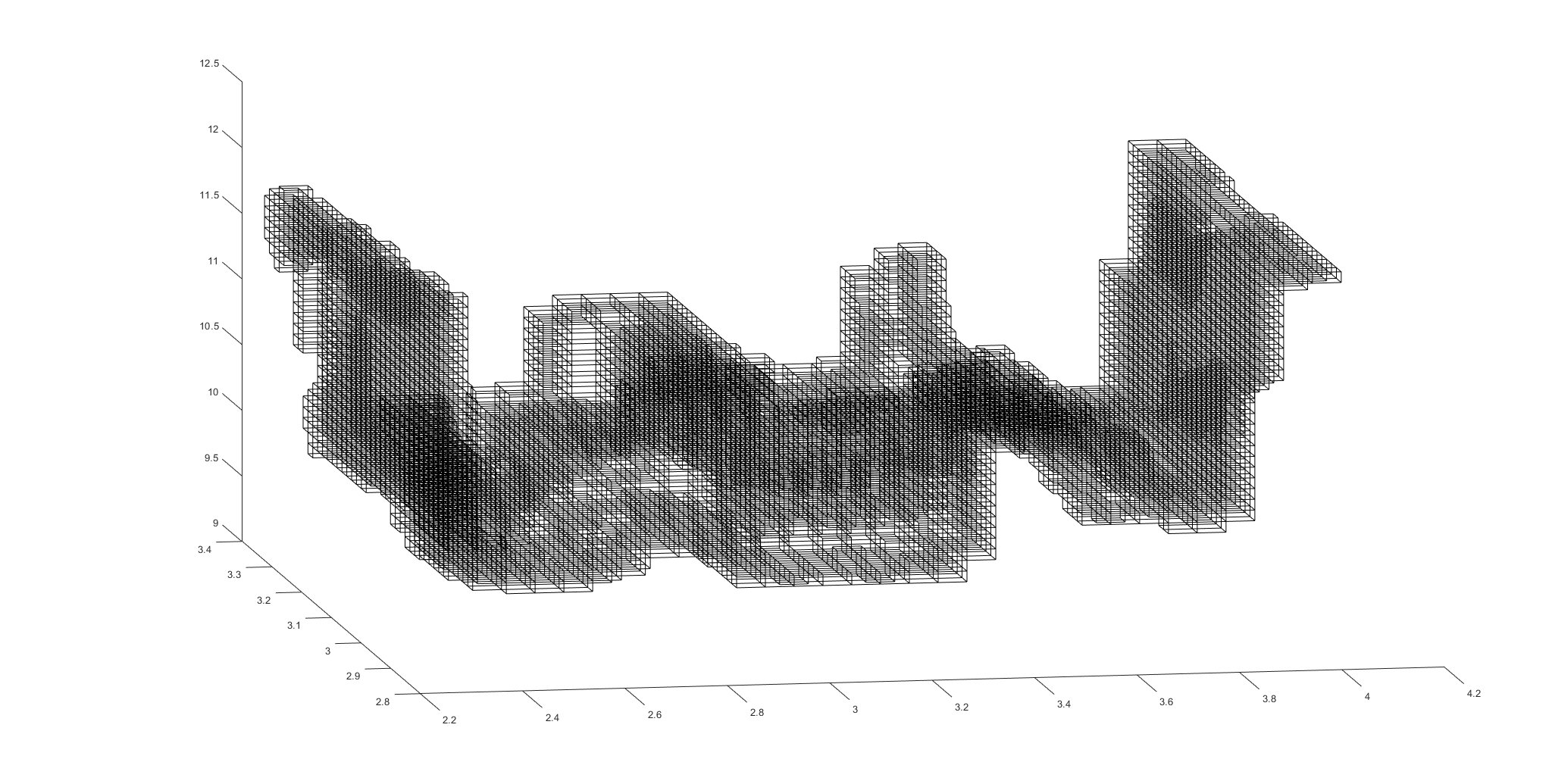}}\
\subfloat[]{\includegraphics[width=0.45\linewidth]{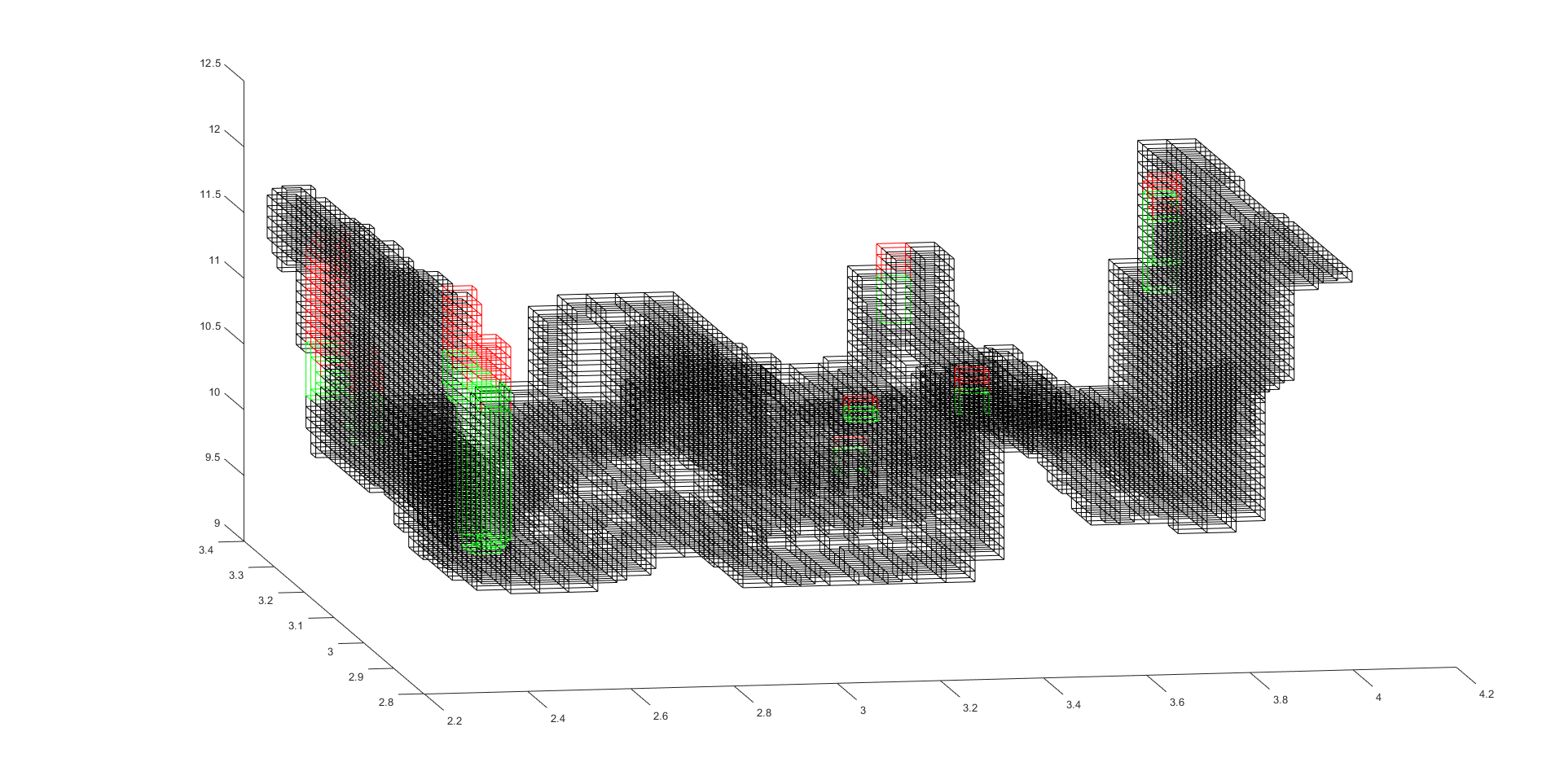}}

\caption{Data set 1. 
A - coloured points (presented 5\% of the total number of the points). B 
 - Points without RGB colour information. C 
 - Cuboids of maximal depth containing points. D - Cuboids of maximal depth containing points with the coloured regions of interest.
}
\label{fig}
\end{figure}

\begin{figure}[htbp]
\centering
\subfloat[]{\includegraphics[width=0.45\linewidth]{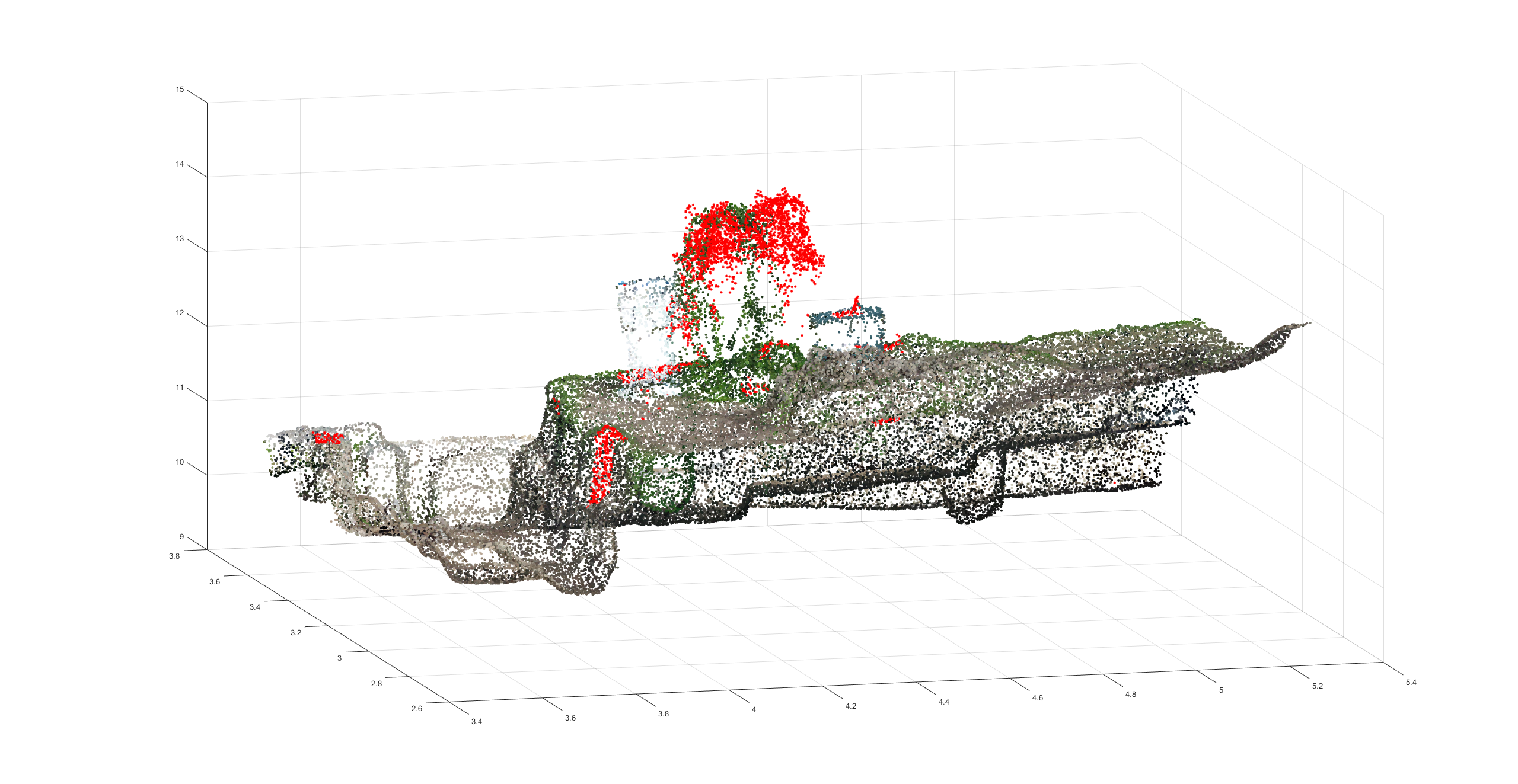}}
\subfloat[]{\includegraphics[width=0.45\linewidth]{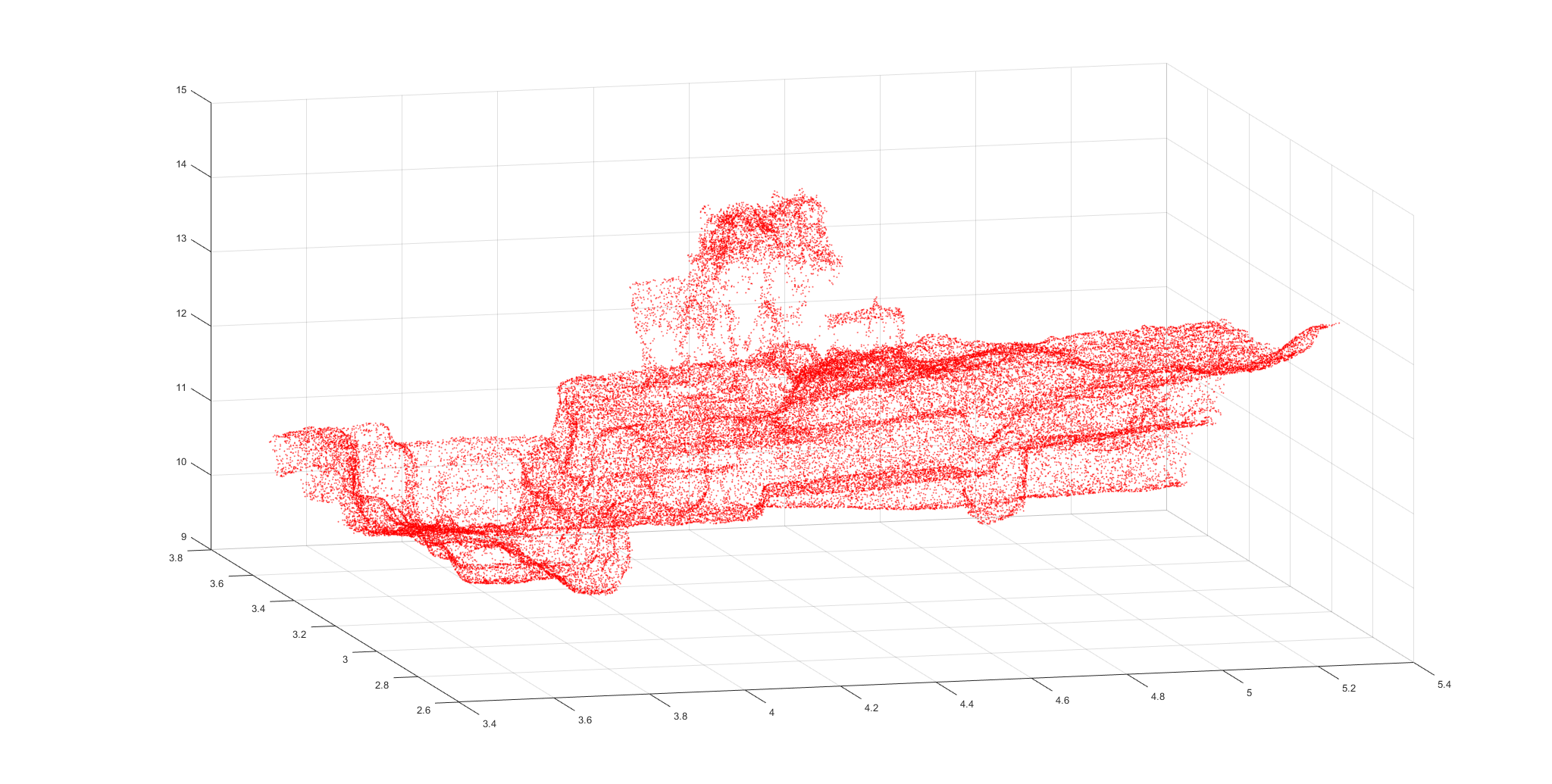}}\\
\subfloat[]{\includegraphics[width=0.45\linewidth]{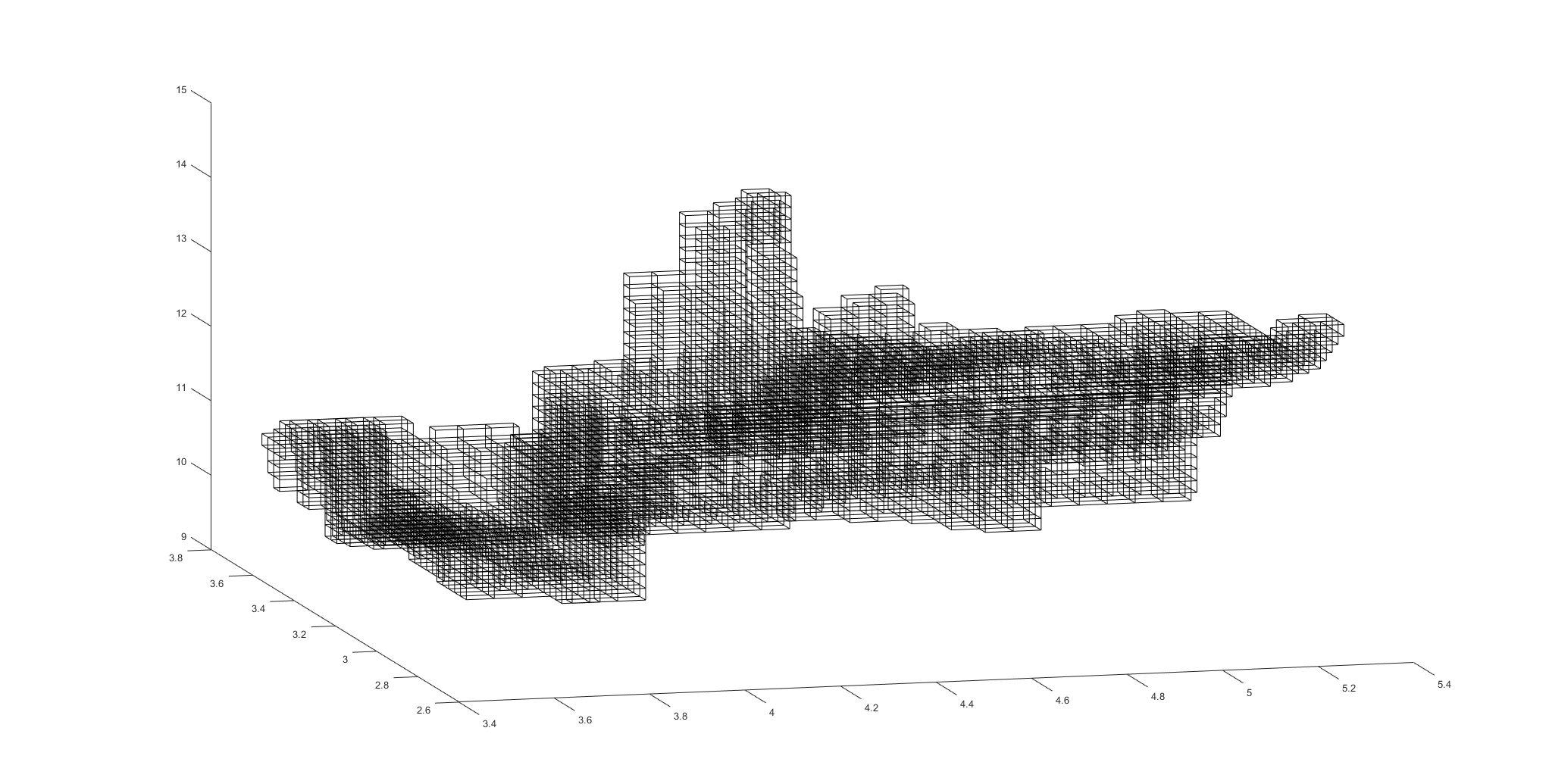}}\
\subfloat[]{\includegraphics[width=0.45\linewidth]{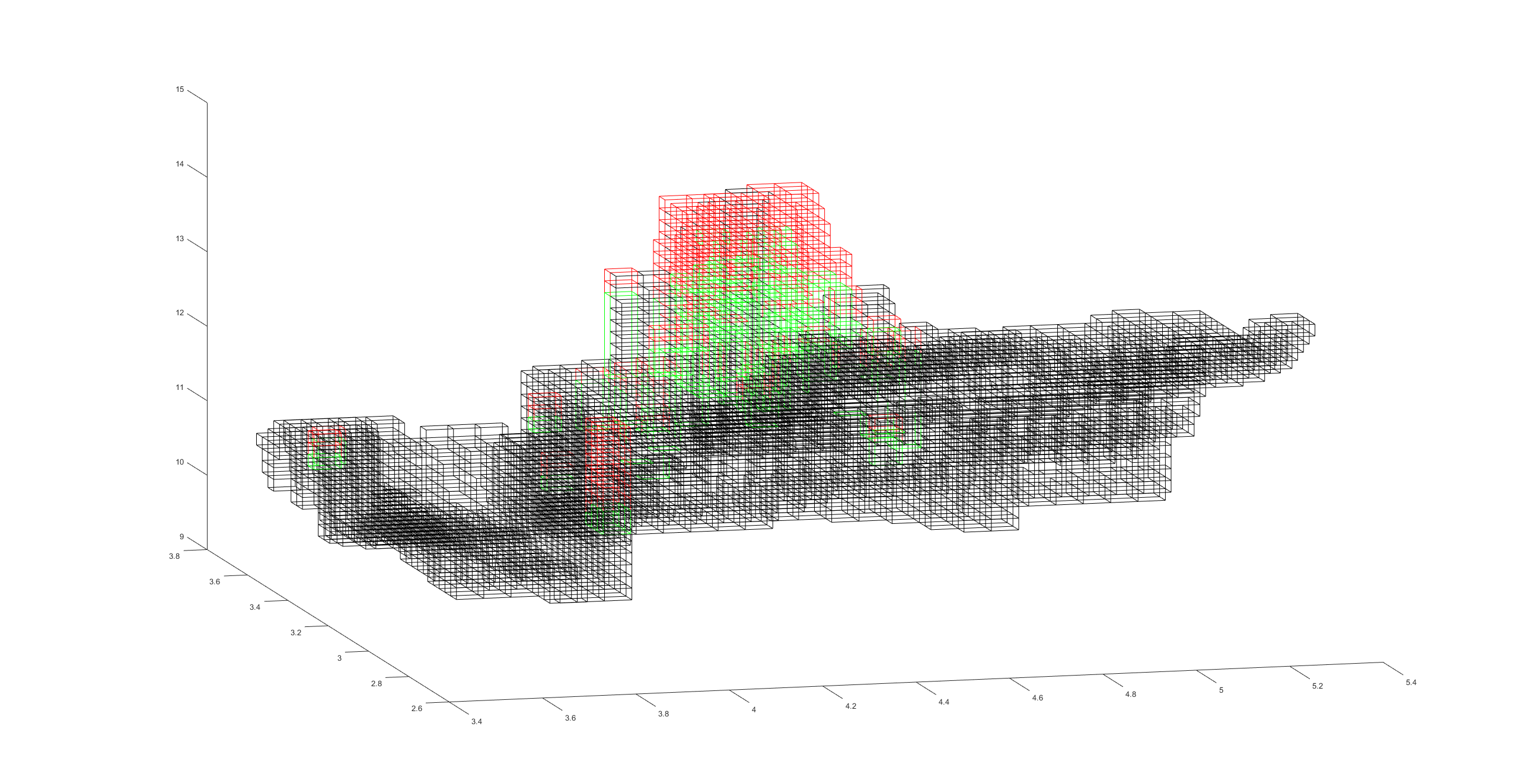}}\

\caption{Data set 2. A - coloured points (presented 5\% of the total number of the points). B - Points without RGB colour information. C - Cuboids of maximal depth containing points. D - Cuboids of maximal depth containing points with the coloured regions of interest.
}
\label{fig2}
\end{figure}

\begin{figure}[htbp]
\centering
\subfloat[]{\includegraphics[width=0.45\linewidth]{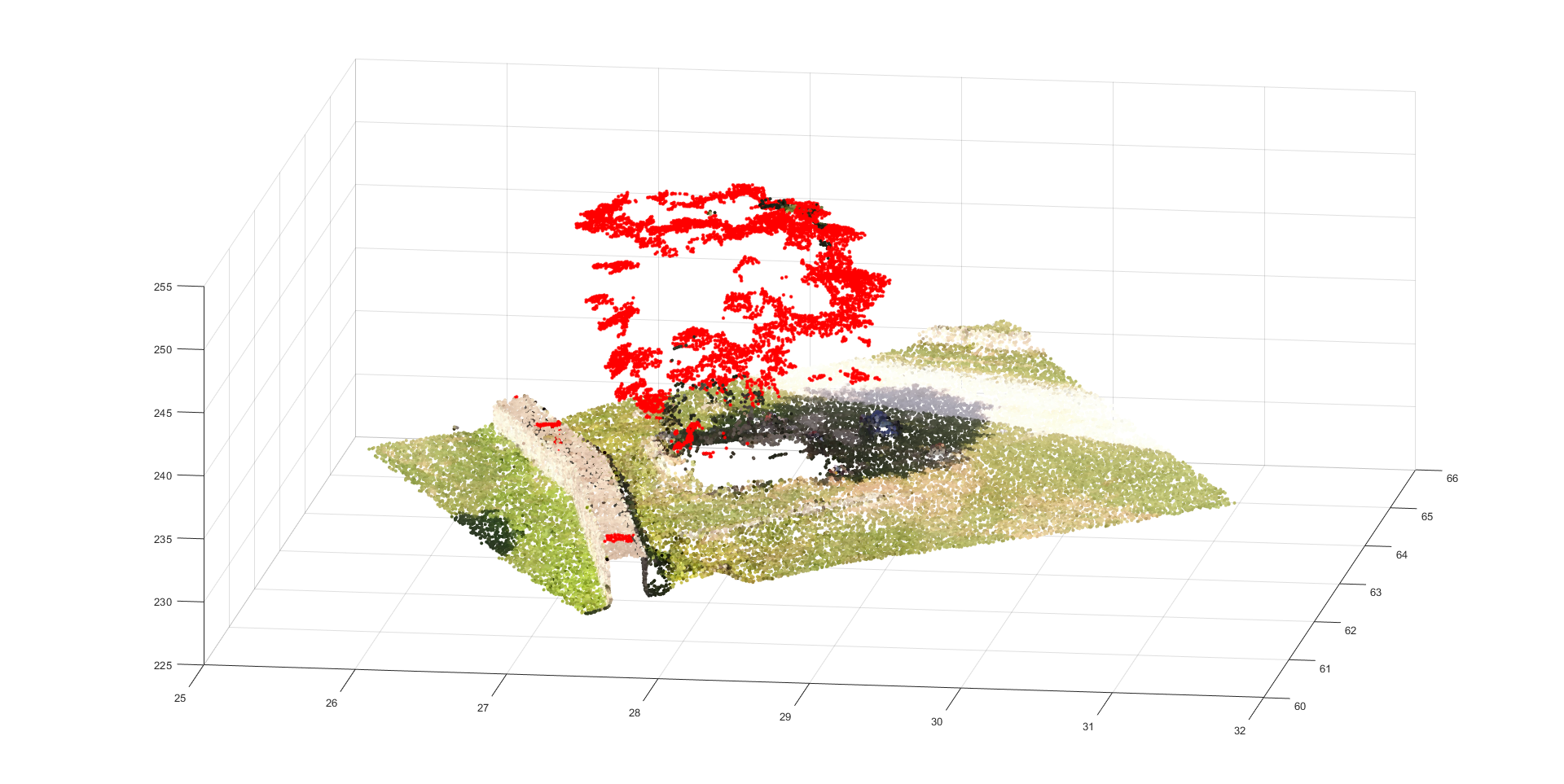}}
\subfloat[]{\includegraphics[width=0.45\linewidth]{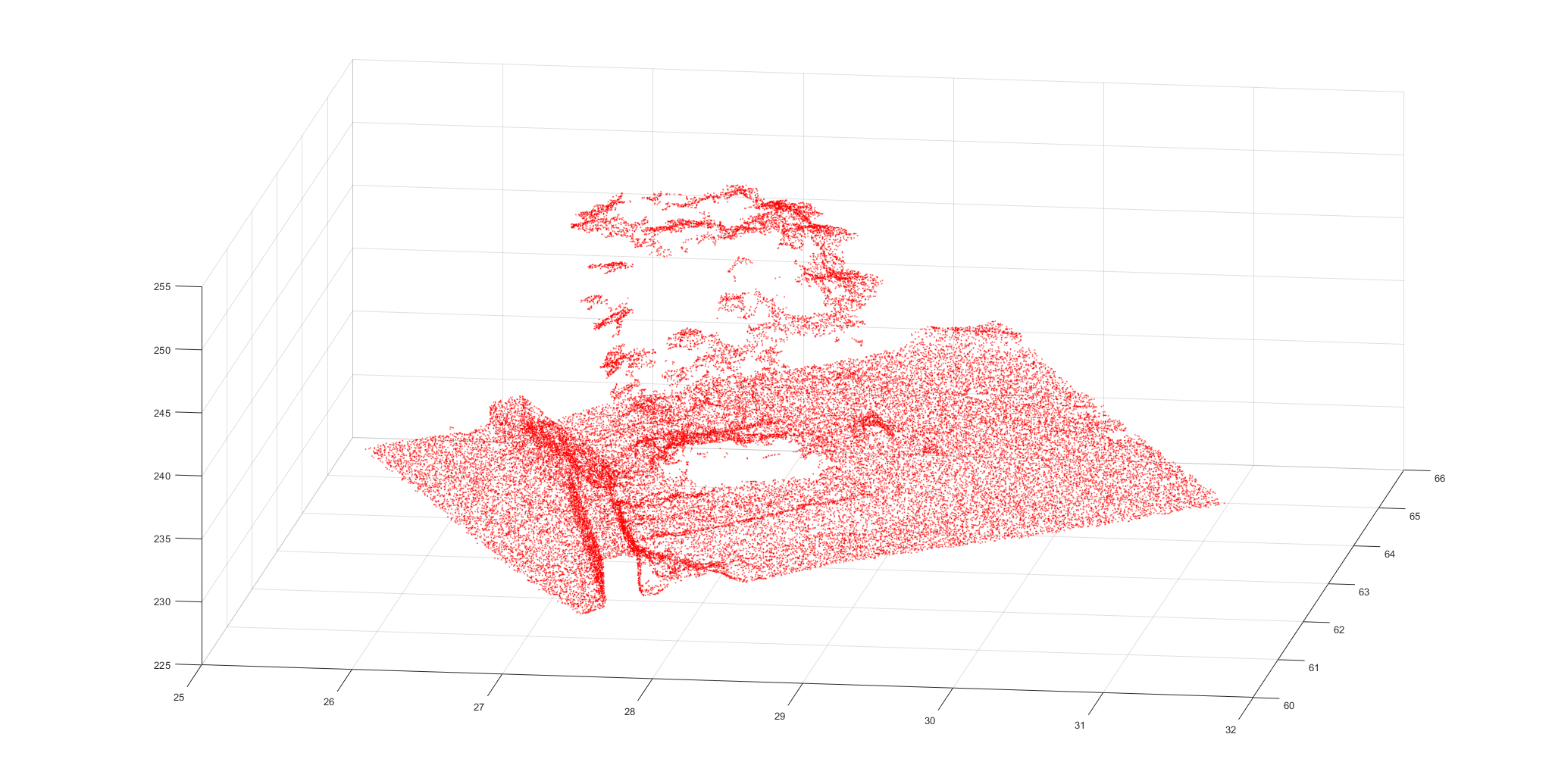}}\\
\subfloat[]{\includegraphics[width=0.45\linewidth]{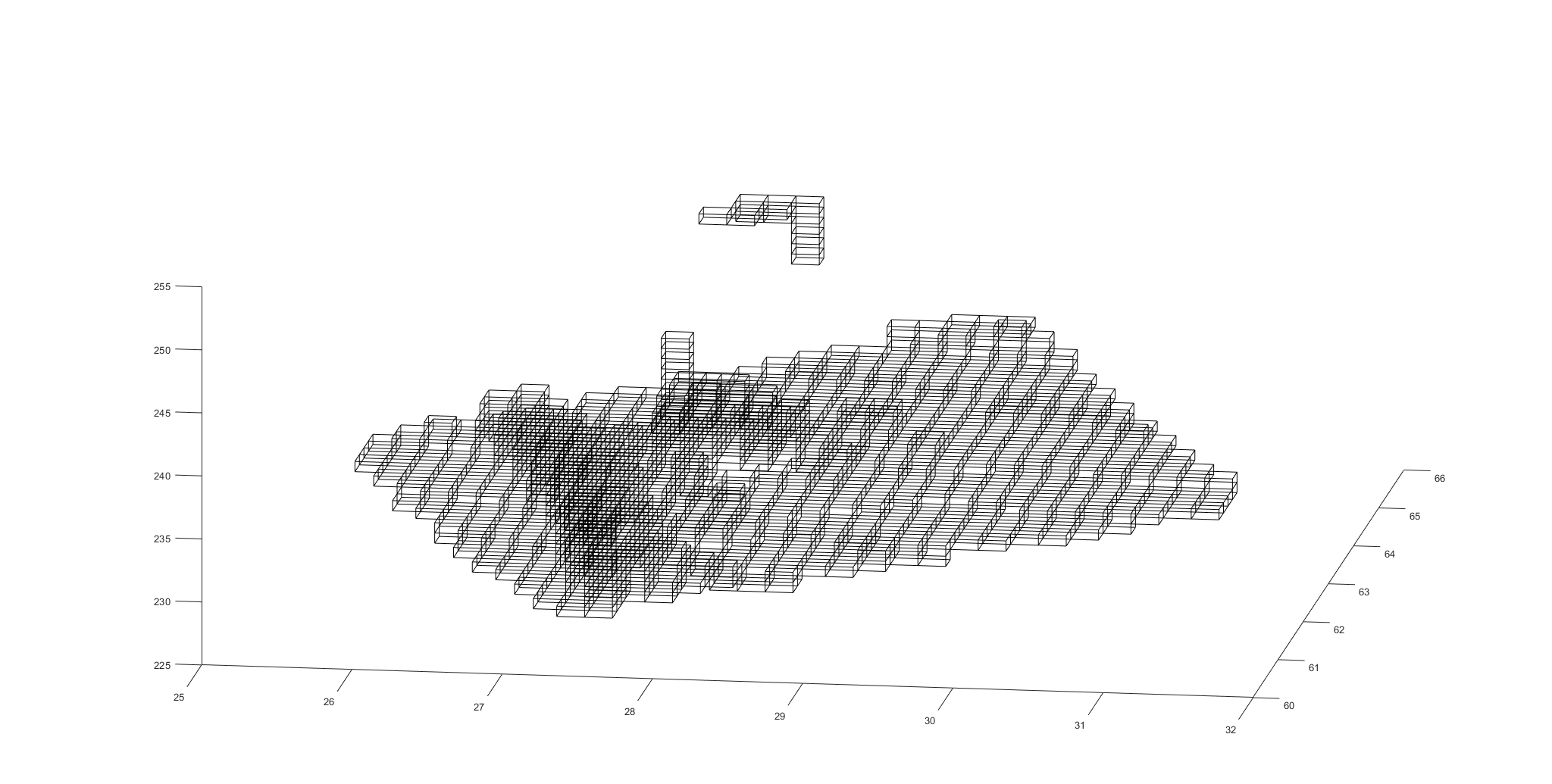}}\
\subfloat[]{\includegraphics[width=0.45\linewidth]{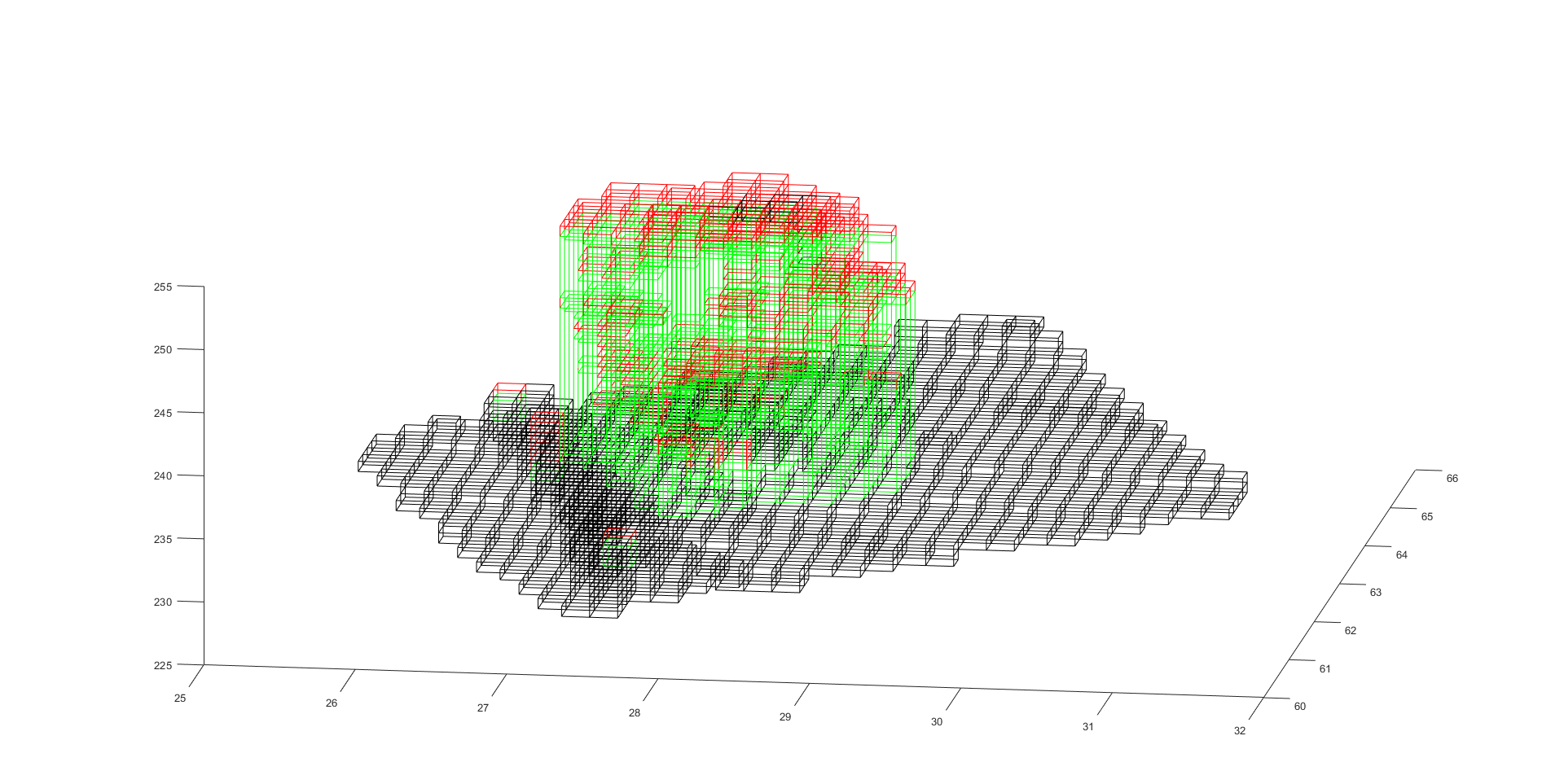}}\

\caption{Data set 3. A - coloured points  (presented 5\% of the total number of the points). B -  Points without RGB colour information. C -  Cuboids of maximal depth containing points. D - Cuboids of maximal depth containing points with the coloured regions of interest.
}
\label{fig3}
\end{figure}

\end{document}